\documentclass{article} % For LaTeX2e
\usepackage{wmw2026_conference,times}

\usepackage{amsmath,amsfonts,bm}

\def\eqref#1{equation~\ref{#1}}
\def\1{\bm{1}}

\DeclareMathAlphabet{\mathsfit}{\encodingdefault}{\sfdefault}{m}{sl}
\SetMathAlphabet{\mathsfit}{bold}{\encodingdefault}{\sfdefault}{bx}{n}

\usepackage{hyperref}
\usepackage{url}
\usepackage{graphicx}
\usepackage{amsmath}
\usepackage{booktabs}
\usepackage{xcolor}
\usepackage{float}

\title{Self-Supervised Multi-Modal World Model with 4D Space-Time Embedding}

\author{Lance Legel\thanks{Correspondence: \texttt{lance@ecodash.ai}} \\
Ecological Intelligence Lab \\
Ecodash.ai \\
\AND
Qin Huang \\
School of Complex Adaptive Systems \\
Arizona State University \\
\And
Brandon Voelker \\
Geosensing Systems Engineering \& Sciences Lab \\
University of Houston \\
\AND
Daniel Neamati \\
Navigation \& Autonomous Vehicles Lab \\
Stanford University \\
\And
Patrick Alan Johnson \\
Earth System Lab \\
Allen Institute for Artificial Intelligence \\
\AND
Favyen Bastani \\
Earth System Lab \\
Allen Institute for Artificial Intelligence \\
\And
Jeff Rose \\
Spatial Intelligence Lab \\
SpatialLogic.com \\
\AND
James Ryan Hennessy \\
Department of Computer Science \\
Georgia Institute of Technology \\
\And
Robert Guralnick \\
Florida Museum of Natural History \\
University of Florida \\
\AND
Douglas Soltis \\
Florida Museum of Natural History \\
University of Florida \\
\And
Pamela Soltis \\
Florida Museum of Natural History \\
University of Florida \\
\AND
Shaowen Wang \\
NSF Institute for Geospatial Understanding \\
University of Illinois Urbana-Champaign \\
}

\wmwfinalcopy % Uncomment for camera-ready version, but NOT for submission.

\begin{document}

\maketitle

\begin{abstract}
We present \textit{DeepEarth}, a self-supervised multi-modal world model with \textit{Earth4D}, a novel planetary-scale 4D space-time positional encoder. Earth4D extends 3D multi-resolution hash encoding to include time, efficiently scaling across the planet over centuries with sub-meter, sub-second precision. Multi-modal encoders (\textit{e.g.} vision-language models) are fused with Earth4D embeddings and trained via masked reconstruction. We demonstrate Earth4D's expressive power by achieving state-of-the-art performance on an ecological forecasting benchmark. Earth4D with learnable hash probing surpasses a multi-modal foundation model pre-trained on substantially more data. Access open source code and download models at: \\
\url{https://github.com/legel/deepearth}.
\end{abstract}

\section{DeepEarth Architecture}
\label{sec:deepearth_architecture}

\begin{figure}[H]
\centering
\includegraphics[width=\textwidth]{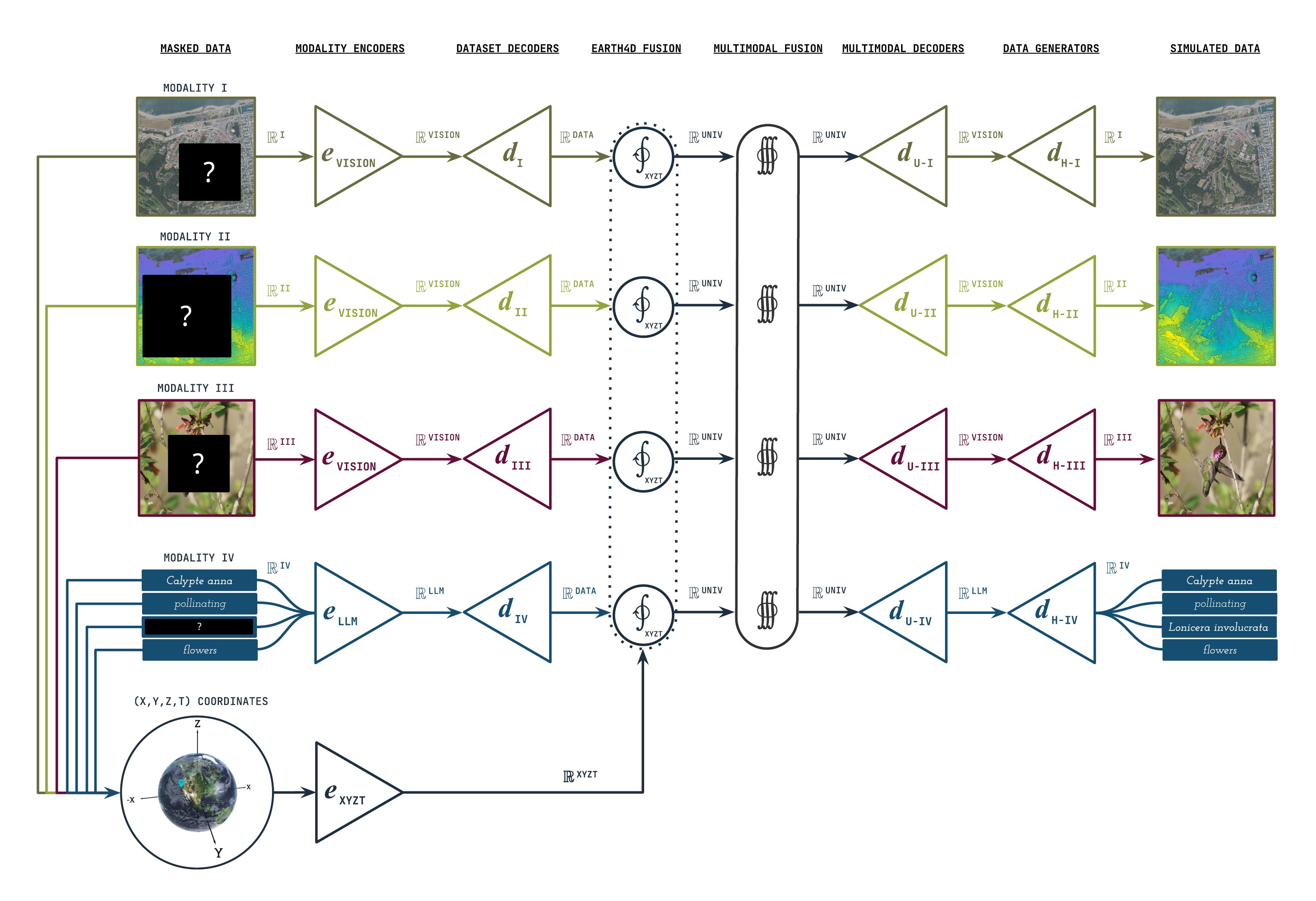}
\caption{\textbf{DeepEarth Overview.} Masked multi-modal data (\textit{e.g.} images, text) sampled around an event (\textit{e.g.} pollination) are encoded and fused with Earth4D space-time embeddings. These universal tokens are jointly encoded, and then masked data is inductively decoded and simulated.}
\label{fig:inductive_simulator}
\end{figure}

DeepEarth is a self-supervised multi-modal world model that learns unified representations of Earth observation data across space and time. As seen in Figure~\ref{fig:inductive_simulator}, the architecture processes multi-modal inputs (\textit{e.g.} vision, language, sensor data) sampled around spatio-temporal events. The Earth4D encoder maps continuous space-time coordinates (\textit{latitude}, \textit{longitude}, \textit{elevation}, \textit{time}) to learnable positional embeddings, which are fused with embeddings from modality-specific encoders and processed as tokens in an autoencoder context window. Inspired by PerceiverIO \citep{Jaegle2021PerceiverIA}, V-JEPA 2 \citep{assran2025vjepa2selfsupervisedvideo}, Galileo \citep{tsenggalileo}, and AlphaEarth \citep{brown2025alphaearthfoundationsembeddingfield}, DeepEarth learns to generatively reconstruct and simulate joint distributions of multi-modal data.

\section{Earth4D Architecture}
\label{sec:earth4d_architecture}

\begin{figure}[H]
  \centering
  \includegraphics[width=\textwidth]{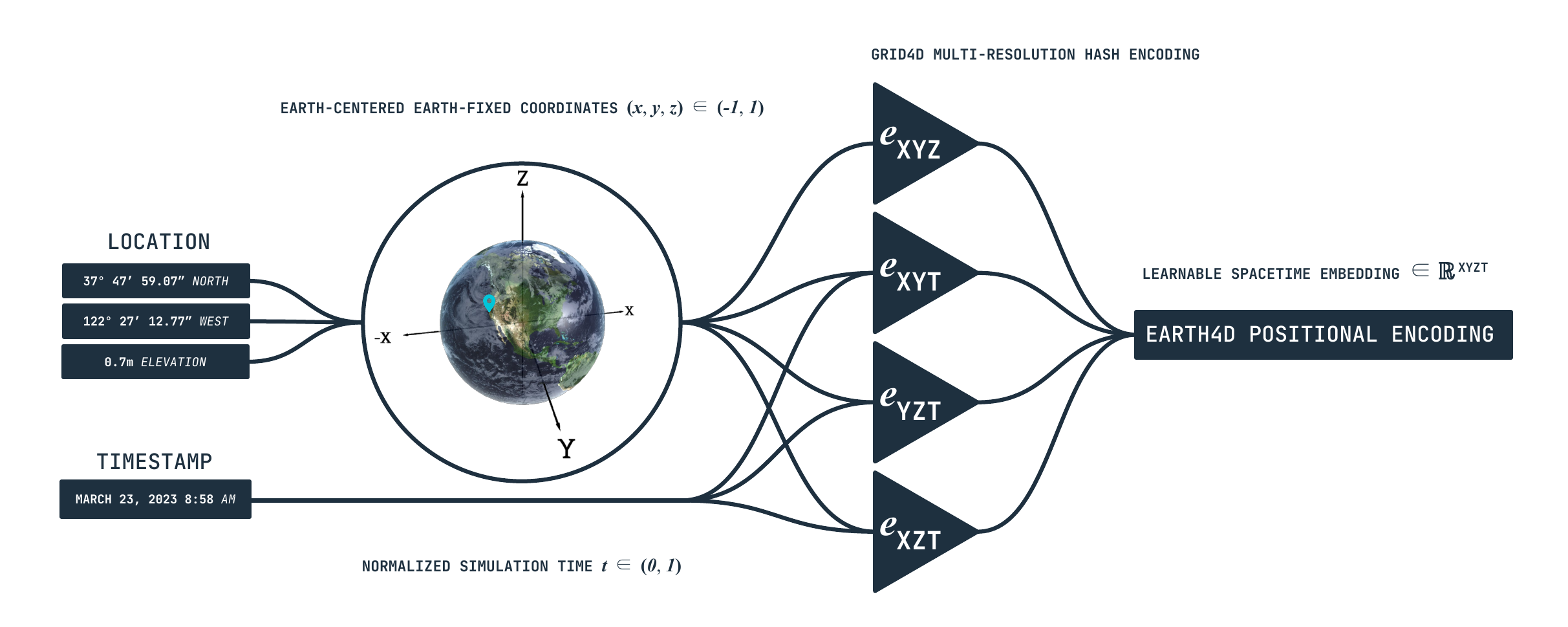}
  \caption{\textbf{Earth4D Space-Time Positional Encoding.} A planetary-scale 4D encoder with fully decomposable spatio-temporal representation. Four grids (\textit{xyz}, \textit{xyt}, \textit{yzt}, \textit{xzt}) are each learned in 3D space and computed in parallel. Each grid has multiple resolution levels (Appendix~\ref{appendix:resolution}), enabling deep learning of complex joint distributions in multi-modal data across space-time scales.
}
  \label{fig:earth4d_encoder}
\end{figure}

Following Grid4D \citep{xu2024grid4d}, Earth4D extends NVIDIA's multi-resolution hash encoding \citep{M_ller_2022} to four dimensions (Figure~\ref{fig:earth4d_encoder}) by concatenating features from one spatial (\textit{xyz}) and three spatio-temporal (\textit{xyt}, \textit{yzt}, \textit{xzt}) grids. Implemented as a standalone PyTorch module with massively parallelizable CUDA kernels, Earth4D is suitable for integration into larger models.

Earth4D's hash encoding compresses spatial features into a fixed memory budget, but different coordinates can map to the same memory location (collisions). We integrate learned hash probing \citep{takikawa2023compactneuralgraphicsprimitives}, an end-to-end differentiable system that learns optimal memory allocation patterns for the data. This yields substantial performance improvements across tasks (Appendix~\ref{appendix:learned_probing}).

\section{Earth4D Experimental Validation}
\label{sec:experiments}

\subsection{Live Fuel Moisture Content Prediction}

\noindent\textbf{\textit{Dataset.}} Live Fuel Moisture Content (LFMC) measures the percentage of water in vegetation relative to its dry weight, a critical indicator for wildfire risk assessment. We evaluate Earth4D on Globe-LFMC 2.0 \citep{yebra2024globe}, a global ecological forecasting benchmark containing field measurements across diverse plant species, geographic regions, and temporal periods.

\noindent\textbf{\textit{Baseline Model.}} We compare against Galileo \citep{johnson2025highresolutionlivefuelmoisture,tsenggalileo}, a pre-trained Vision Transformer processing multi-modal remote sensing data (Appendix~\ref{appendix:benchmark}).

\noindent\textbf{\textit{Architecture.}} Earth4D encodes (\textit{x,y,z,t}) into a 192D vector, concatenated with a learnable species embedding initialized randomly (no prior knowledge). An MLP then predicts LFMC \%.

\begin{figure}[t]
\centering
\includegraphics[width=\textwidth]{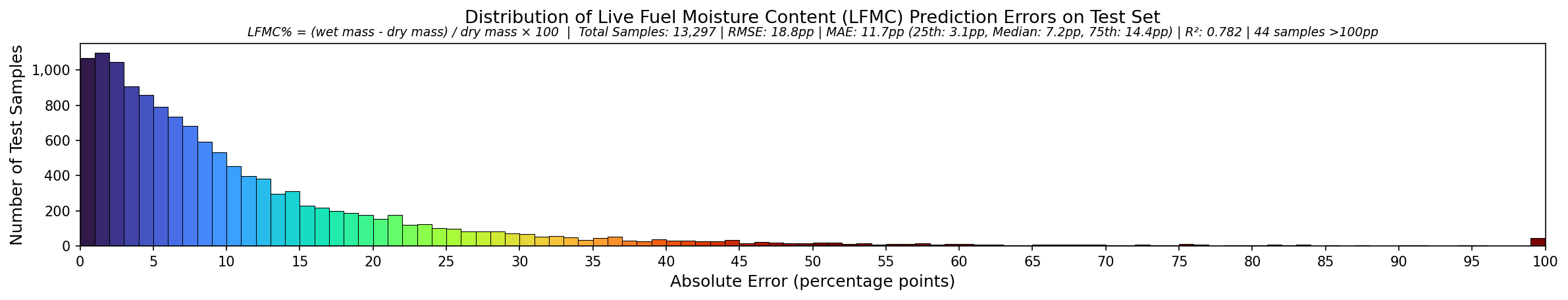}
\includegraphics[width=\textwidth]{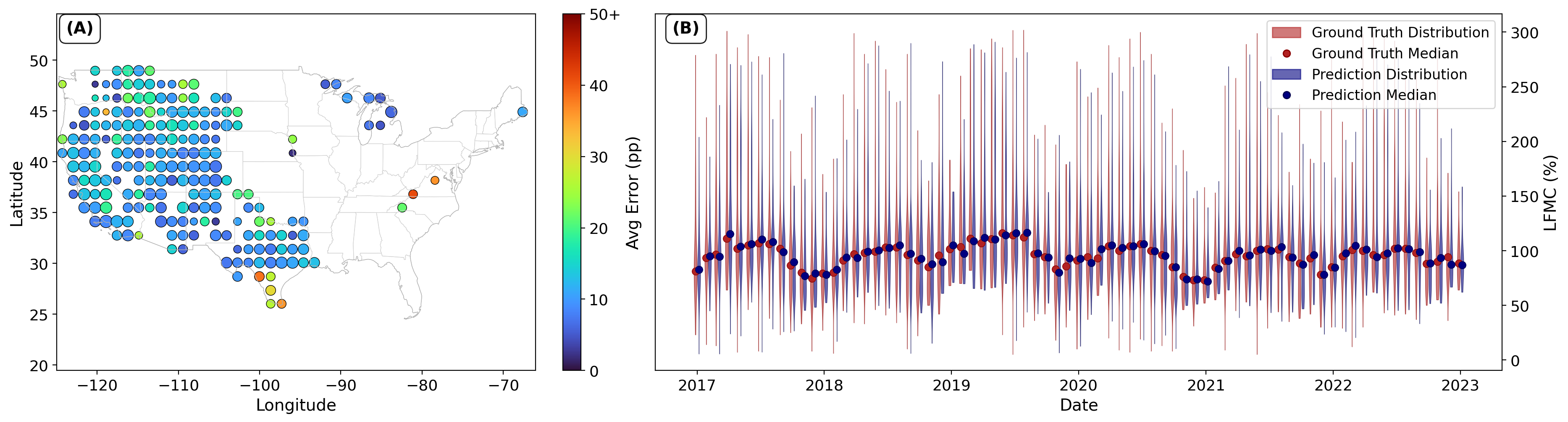}
\caption{\textbf{Earth4D LFMC Prediction Performance.} \textit{(Top)} Distribution of absolute errors in percentage point predictions across 13,297 test samples, showing median error of 7.1pp. \textit{(A)} Geographic error distribution across CONUS shows low error in well-sampled regions. \textit{(B)} Temporal predictions closely track ground truth LFMC measurements across seasons (2017--2023).}
\label{fig:error_dist}
\end{figure}

\noindent\textbf{\textit{Results.}} Earth4D achieves MAE 11.7pp and R² 0.783, surpassing Galileo (MAE 12.6pp, R² 0.72) using only (\textit{x,y,z,t}) coordinates and species embeddings (Table~\ref{tab:lfmc_results}).

\begin{table}[h]
\centering
\small
\begin{tabular}{@{}llccc@{}}
\toprule
\textbf{Model} & \textbf{Data Inputs} & \textbf{MAE (pp)} & \textbf{RMSE (pp)} & \textbf{R²} \\
\midrule
\textbf{\textit{Galileo}} (Pre-Trained) & (\textit{x,y,z,t}) + Species Type + Remote Sensing & 12.6 & 18.9 & 0.72 \\
\textbf{\textit{Earth4D}} (Learned Hashing) & (\textit{x,y,z,t}) + Species Name & \textbf{11.7} & \textbf{18.7} & \textbf{0.783} \\
\bottomrule
\end{tabular}
\caption{\textbf{State-of-the-Art Ecological Forecasting Benchmark.} Earth4D surpasses the pre-trained Galileo foundation model without satellite imagery, weather data, or topography.}
\label{tab:lfmc_results}
\end{table}

% \subsection{RGB Aerial Imagery Reconstruction}
%
% We evaluate Earth4D's ability to infer RGB pixels from (\textit{x,y,z,t}) inputs with objective (\textit{x,y,z,t}) $\rightarrow$ (\textit{r,g,b}). Using USGS 3DEP LiDAR \citep{stoker2022accuracy,sugarbaker20143d} and USDA NAIP imagery \citep{USDA_NAIP} paired by \citet{allred2025canopy}, we train on 5.8M coordinate-color pairs from Houston coastal wetlands (Figure~\ref{fig:rgb_reconstruction}).
%
% \begin{figure}[H]
% \centering
% \includegraphics[width=\textwidth]{figures/rgb_reconstruction.png}
% \caption{\textbf{RGB Reconstruction from LiDAR Elevation.} Houston coastal wetlands, 2018. \textit{Left to right:} LiDAR height, ground truth, baseline, learned probing (18\% lower loss).}
% \label{fig:rgb_reconstruction}
% \end{figure}

\newpage

\bibliography{deepearth}

\begin{thebibliography}{15}
\providecommand{\natexlab}[1]{#1}
\providecommand{\url}[1]{\texttt{#1}}
\expandafter\ifx\csname urlstyle\endcsname\relax
  \providecommand{\doi}[1]{doi: #1}\else
  \providecommand{\doi}{doi: \begingroup \urlstyle{rm}\Url}\fi

\bibitem[Abatzoglou et~al.(2018)Abatzoglou, Dobrowski, Parks, and
  Hegewisch]{abatzoglou2018terraclimate}
John~T Abatzoglou, Solomon~Z Dobrowski, Sean~A Parks, and Katherine~C
  Hegewisch.
\newblock {TerraClimate}, a high-resolution global dataset of monthly climate
  and climatic water balance from 1958--2015.
\newblock \emph{Scientific Data}, 5:\penalty0 1--12, 2018.

\bibitem[Assran et~al.(2025)Assran, Bardes, Fan, Garrido, Howes, Mojtaba,
  Komeili, Muckley, Rizvi, Roberts, Sinha, Zholus, Arnaud, Gejji, Martin,
  Hogan, Dugas, Bojanowski, Khalidov, Labatut, Massa, Szafraniec, Krishnakumar,
  Li, Ma, Chandar, Meier, LeCun, Rabbat, and
  Ballas]{assran2025vjepa2selfsupervisedvideo}
Mido Assran, Adrien Bardes, David Fan, Quentin Garrido, Russell Howes, Mojtaba,
  Komeili, Matthew Muckley, Ammar Rizvi, Claire Roberts, Koustuv Sinha, Artem
  Zholus, Sergio Arnaud, Abha Gejji, Ada Martin, Francois~Robert Hogan, Daniel
  Dugas, Piotr Bojanowski, Vasil Khalidov, Patrick Labatut, Francisco Massa,
  Marc Szafraniec, Kapil Krishnakumar, Yong Li, Xiaodong Ma, Sarath Chandar,
  Franziska Meier, Yann LeCun, Michael Rabbat, and Nicolas Ballas.
\newblock {V-JEPA} 2: Self-supervised video models enable understanding,
  prediction and planning, 2025.
\newblock URL \url{https://arxiv.org/abs/2506.09985}.

\bibitem[Brown et~al.(2025)Brown, Kazmierski, Pasquarella, Rucklidge,
  Samsikova, Zhang, Shelhamer, Lahera, Wiles, Ilyushchenko, Gorelick, Zhang,
  Alj, Schechter, Askay, Guinan, Moore, Boukouvalas, and
  Kohli]{brown2025alphaearthfoundationsembeddingfield}
Christopher~F. Brown, Michal~R. Kazmierski, Valerie~J. Pasquarella, William~J.
  Rucklidge, Masha Samsikova, Chenhui Zhang, Evan Shelhamer, Estefania Lahera,
  Olivia Wiles, Simon Ilyushchenko, Noel Gorelick, Lihui~Lydia Zhang, Sophia
  Alj, Emily Schechter, Sean Askay, Oliver Guinan, Rebecca Moore, Alexis
  Boukouvalas, and Pushmeet Kohli.
\newblock {AlphaEarth Foundations}: An embedding field model for accurate and
  efficient global mapping from sparse label data, 2025.
\newblock URL \url{https://arxiv.org/abs/2507.22291}.

\bibitem[Dosovitskiy et~al.(2021)Dosovitskiy, Beyer, Kolesnikov, Weissenborn,
  Zhai, Unterthiner, Dehghani, Minderer, Heigold, Gelly, Uszkoreit, and
  Houlsby]{dosovitskiy2021an}
Alexey Dosovitskiy, Lucas Beyer, Alexander Kolesnikov, Dirk Weissenborn,
  Xiaohua Zhai, Thomas Unterthiner, Mostafa Dehghani, Matthias Minderer, Georg
  Heigold, Sylvain Gelly, Jakob Uszkoreit, and Neil Houlsby.
\newblock An image is worth 16x16 words: Transformers for image recognition at
  scale.
\newblock In \emph{International Conference on Learning Representations}, 2021.
\newblock URL \url{https://openreview.net/forum?id=YicbFdNTTy}.

\bibitem[Drusch et~al.(2012)Drusch, Del~Bello, Carlier, Colin, Fernandez,
  Gascon, Hoersch, Isola, Laberinti, Martimort, et~al.]{drusch2012sentinel}
Matthias Drusch, Umberto Del~Bello, S{\'e}bastien Carlier, Olivier Colin,
  Veronica Fernandez, Ferran Gascon, Bianca Hoersch, Claudia Isola, Paolo
  Laberinti, Philippe Martimort, et~al.
\newblock {Sentinel-2}: {ESA's} optical high-resolution mission for {GMES}
  operational services.
\newblock \emph{Remote Sensing of Environment}, 120:\penalty0 25--36, 2012.

\bibitem[Farr \& Kobrick(2000)Farr and Kobrick]{farr2000shuttle}
Tom~G Farr and Mike Kobrick.
\newblock Shuttle radar topography mission produces a wealth of data.
\newblock \emph{Eos, Transactions American Geophysical Union}, 81:\penalty0
  583--585, 2000.

\bibitem[Jaegle et~al.()Jaegle, Borgeaud, Alayrac, Doersch, Ionescu, Ding,
  Koppula, Zoran, Brock, Shelhamer, et~al.]{Jaegle2021PerceiverIA}
Andrew Jaegle, Sebastian Borgeaud, Jean-Baptiste Alayrac, Carl Doersch, Catalin
  Ionescu, David Ding, Skanda Koppula, Daniel Zoran, Andrew Brock, Evan
  Shelhamer, et~al.
\newblock Perceiver {IO}: A general architecture for structured inputs \&
  outputs.
\newblock In \emph{International Conference on Learning Representations}.

\bibitem[Jiawei et~al.(2024)Jiawei, Zexin, Jian, and Jin]{xu2024grid4d}
Xu~Jiawei, Fan Zexin, Yang Jian, and Xie Jin.
\newblock {Grid4D}: {4D} decomposed hash encoding for high-fidelity dynamic
  gaussian splatting.
\newblock \emph{The Thirty-eighth Annual Conference on Neural Information
  Processing Systems}, 2024.

\bibitem[Johnson et~al.(2025)Johnson, Tseng, Zhang, Heward, Sjahli, Bastani,
  Redmon, and Beukema]{johnson2025highresolutionlivefuelmoisture}
Patrick~Alan Johnson, Gabriel Tseng, Yawen Zhang, Heather Heward, Virginia
  Sjahli, Favyen Bastani, Joseph Redmon, and Patrick Beukema.
\newblock High-resolution live fuel moisture content ({LFMC}) maps for wildfire
  risk from multimodal earth observation data, 2025.
\newblock URL \url{https://arxiv.org/abs/2506.20132}.

\bibitem[Mu{\~n}oz~Sabater(2019)]{munozsabater2019era5}
Joaqu{\'i}n Mu{\~n}oz~Sabater.
\newblock {ERA5-Land} monthly averaged data from 1950 to present.
\newblock Copernicus Climate Change Service (C3S) Climate Data Store (CDS),
  2019.

\bibitem[Müller et~al.(2022)Müller, Evans, Schied, and Keller]{M_ller_2022}
Thomas Müller, Alex Evans, Christoph Schied, and Alexander Keller.
\newblock Instant neural graphics primitives with a multiresolution hash
  encoding.
\newblock \emph{ACM Transactions on Graphics}, 41\penalty0 (4):\penalty0
  1–15, July 2022.
\newblock ISSN 1557-7368.
\newblock \doi{10.1145/3528223.3530127}.
\newblock URL \url{http://dx.doi.org/10.1145/3528223.3530127}.

\bibitem[Takikawa et~al.(2023)Takikawa, Müller, Nimier-David, Evans, Fidler,
  Jacobson, and Keller]{takikawa2023compactneuralgraphicsprimitives}
Towaki Takikawa, Thomas Müller, Merlin Nimier-David, Alex Evans, Sanja Fidler,
  Alec Jacobson, and Alexander Keller.
\newblock Compact neural graphics primitives with learned hash probing, 2023.
\newblock URL \url{https://arxiv.org/abs/2312.17241}.

\bibitem[Torres et~al.(2012)Torres, Snoeij, Geudtner, Bibby, Davidson, Attema,
  Potin, Rommen, Floury, Brown, et~al.]{torres2012gmes}
Ramon Torres, Paul Snoeij, Dirk Geudtner, David Bibby, Malcolm Davidson, Evert
  Attema, Pierre Potin, B{\"o}rn Rommen, Nicolas Floury, Mike Brown, et~al.
\newblock {GMES} {Sentinel-1} mission.
\newblock \emph{Remote Sensing of Environment}, 120:\penalty0 9--24, 2012.

\bibitem[Tseng et~al.(2025)Tseng, Fuller, Reil, Herzog, Beukema, Bastani,
  Green, Shelhamer, Kerner, and Rolnick]{tsenggalileo}
Gabriel Tseng, Anthony Fuller, Marlena Reil, Henry Herzog, Patrick Beukema,
  Favyen Bastani, James~R Green, Evan Shelhamer, Hannah Kerner, and David
  Rolnick.
\newblock {Galileo}: Learning global \& local features of many remote sensing
  modalities.
\newblock In \emph{Forty-second International Conference on Machine Learning},
  2025.

\bibitem[Yebra et~al.(2024)Yebra, Scortechini, Adeline, Aktepe, Almoustafa,
  Bar-Massada, Beget, Boer, Bradstock, Brown, et~al.]{yebra2024globe}
Marta Yebra, Gianluca Scortechini, Karine Adeline, Nursema Aktepe, Turkia
  Almoustafa, Avi Bar-Massada, Mar{\'\i}a~Eugenia Beget, Matthias Boer, Ross
  Bradstock, Tegan Brown, et~al.
\newblock {Globe-LFMC} 2.0, an enhanced and updated dataset for live fuel
  moisture content research.
\newblock \emph{Scientific Data}, 11\penalty0 (1):\penalty0 332, 2024.

\end{thebibliography}
\bibliographystyle{wmw2026_conference}

\newpage

\appendix

\section*{APPENDICES}
\addcontentsline{toc}{section}{Appendices}

\section{Earth4D Resolution Specifications}
\label{appendix:resolution}

\begin{figure}[H]
\centering
\includegraphics[width=\textwidth]{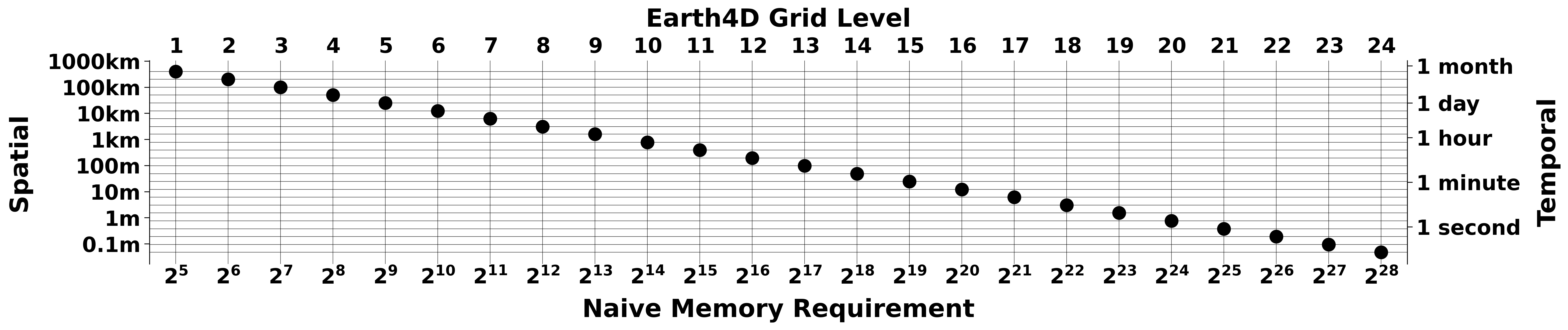}
\caption{\textbf{Earth4D Space-Time Scales.} Default 24$\times$24$\times$24 levels for each \textit{xyz}, \textit{xyt}, \textit{yzt}, \textit{xzt} grid. Each level stores up to $2^{22}$ entries, with each entry storing a 2D feature. Requires 724M trainable parameters ($\sim$11 GB GPU memory during training). Parallelizable across levels and spatio-temporal boundaries. Outputs 192D per $(x,y,z,t)$ coordinate from 4 grids $\times$ 24 levels $\times$ 2D feature per level. Hashing saves memory vs.\ naive requirement, e.g., $(2^{28})^3 = 10^{25}$ at level 24.}
\label{fig:earth4d_specs}
\end{figure}

\newpage

\section{Learned Hash Probing and Ablation Studies}
\label{appendix:learned_probing}

\subsection{Hash Collision Simulations}
\label{appendix:collisions}

\begin{figure}[H]
\centering
\footnotesize
\setlength{\tabcolsep}{4pt}
\begin{tabular}{@{}llll@{}}
\toprule
\textbf{Scenario} & \textbf{Spatial} & \textbf{Temporal} & \textbf{Description} \\
\midrule
\textit{Uniform Random} & Global & Full & Uniform Earth surface sampling \\
\textit{Continental Sparse} & North America & Full & Sparse continental coverage \\
\textit{Moderate Spatial Cluster} & 10km $\times$ 10km & Full & City-scale clustering \\
\textit{Moderate Temporal Cluster} & 1k locations & Distributed & Temporal sampling at fixed locations \\
\textit{Moderate Spatiotemporal} & 1km $\times$ 1km & 1 hour & Neighborhood-scale event \\
\textit{Extreme Spatial Single} & 10m $\times$ 10m & Full & Building-scale dense clustering \\
\textit{Extreme Spatial Multi} & 10 $\times$ (10m $\times$ 10m) & Full & 10 dense clusters worldwide \\
\textit{Extreme Temporal Single} & Global & 1 hour & Global snapshot \\
\textit{Extreme Temporal Multi} & Global & 10 $\times$ (1 hour) & 10 temporal snapshots \\
\textit{Time Series} & 10k locations & 100 steps & Regular temporal sampling \\
\bottomrule
\end{tabular}

\vspace{0.5em}

\includegraphics[width=\textwidth]{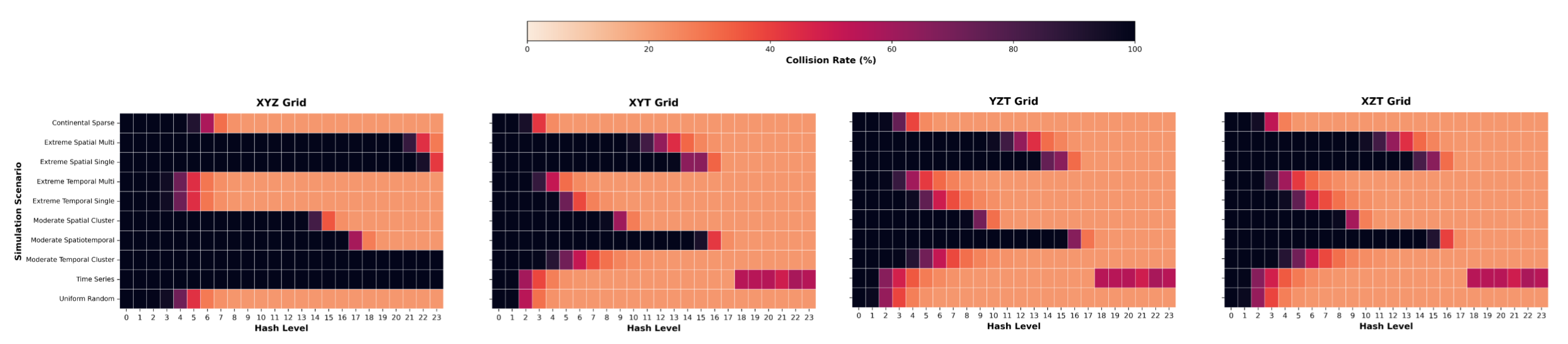}
\caption{\textbf{Earth4D Hash Collision Analysis.} \textit{(Table)} 10 $(x,y,z,t)$ point distribution scenarios that were simulated to analyze hash collisions in Earth4D memory. \textit{(Graph)} Shows results for 1M point simulations across all 24 levels.}
\label{fig:collision_rates}
\end{figure}

\subsection{Performance Improvements}
\label{appendix:baseline_results}

Standard multi-resolution hash encoding without learned probing obtains RMSE 26.0pp, MAE 16.6pp, and R² 0.58 (800M parameters, $2^{22}$ hash capacity). Integrating learned hash probing \citep{takikawa2023compactneuralgraphicsprimitives}, which learns to select optimal hash table indices from a candidate set, yields RMSE 18.7pp, MAE 11.7pp, and R² 0.783—a 29.5\% MAE reduction and 35.0\% R² improvement. Extreme compression to 5M parameters (99.3\% reduction, $2^{14}$ hash capacity) achieves MAE 15.0pp/R² 0.668, outperforming the 800M baseline by 14.7\% in R² with 4$\times$ training speedup and 93\% memory reduction. On RGB reconstruction, learned probing reduces validation loss by 18\%. These gains result from collision reduction (33\% at 1M points) and learned shared features across memory indices, allowing the model to discover meaningful spatio-temporal patterns.

\newpage

\section{Benchmark Specifications}
\label{appendix:benchmark}

\subsection{Galileo Baseline Model}

Galileo \citep{johnson2025highresolutionlivefuelmoisture,tsenggalileo} is a Vision Transformer \citep{dosovitskiy2021an} (5.3M parameters) pre-trained by the Allen Institute for AI. It processes Sentinel-2 optical imagery \citep{drusch2012sentinel}, Sentinel-1 SAR \citep{torres2012gmes}, VIIRS night lights, ERA-5 weather \citep{munozsabater2019era5}, TerraClimate soil/water data \citep{abatzoglou2018terraclimate}, SRTM topography \citep{farr2000shuttle}, (\textit{x,y,z,t}) coordinates, and species type. We use the Allen Institute for AI's exact Globe-LFMC 2.0 \citep{yebra2024globe} train/test split (76,467/13,297) to directly compare against this benchmark.

\end{document}